\def\year{2021}\relax
\documentclass[letterpaper]{article} 
\usepackage{aaai21}  

\usepackage{times}  
\usepackage{helvet} 
\usepackage{courier}  
\usepackage[hyphens]{url}  
\usepackage{graphicx} 
\usepackage{natbib}  
\usepackage{caption} 
\usepackage{latexsym}

\usepackage{microtype}
\usepackage{url}            
\usepackage{booktabs}       
\usepackage{amsfonts}       
\usepackage{nicefrac}       
\usepackage{microtype}      
\usepackage{bbold}
\usepackage{bm}
\usepackage{chngpage}
\usepackage{algorithmic}
\usepackage{algorithm}
\usepackage[algo2e,linesnumbered]{algorithm2e}
\usepackage{graphicx}
\usepackage{epstopdf}
\usepackage{enumitem}
\usepackage{amsmath}
\usepackage{pifont}
\usepackage{amssymb}
\usepackage{pifont}
\usepackage{subfigure}
\usepackage{amsfonts}
\usepackage{mathrsfs}
\usepackage{amsthm}
\usepackage{xcolor}
\nocopyright
\title{Multi-scale Cooperative Multimodal Transformers for Multimodal Sentiment Analysis in Videos}

\author{
    Lianyang Ma\textsuperscript{\rm 1}\thanks{The corresponding author for this paper and the principal investigator for this project is Lianyang Ma.},Yu Yao\textsuperscript{\rm 2},
    Tao Liang\textsuperscript{\rm 1}, Tongliang Liu\textsuperscript{\rm 2}
}
\affiliations{
    \\
    \textsuperscript{\rm 1}Tencent PCG,
    \textsuperscript{\rm 2}University of Sydney\\

    mlysjtu@foxmail.com \qquad \{yyao0814,tongliang.liu\}@sydney.edu.au  \qquad alphaliang@tencent.com
    
}

\begin{document}
\maketitle
\begin{abstract}
Multimodal sentiment analysis in videos is a key task in many real-world applications, which usually requires integrating multimodal streams including visual, verbal and acoustic behaviors.
To improve the robustness of multimodal fusion, some of the existing methods let different modalities communicate with each other and modal the crossmodal interaction via transformers. However, these methods only use the \textit{single-scale} representations during the interaction but forget to exploit \textit{multi-scale} representations that contain different levels of semantic information.
As a result, the representations learned by transformers could be biased especially for \textit{unaligned} multimodal data. 
In this paper, we propose a multi-scale cooperative multimodal transformer (MCMulT) architecture for multimodal sentiment analysis. On the whole, the ``multi-scale'' mechanism is capable of exploiting the different levels of semantic information of each modality which are used for fine-grained crossmodal interactions. Meanwhile, each modality learns its feature hierarchies via integrating the crossmodal interactions from multiple level features of its source modality. In this way, each pair of modalities progressively builds feature hierarchies respectively in a cooperative manner. The empirical results illustrate that our MCMulT model not only outperforms existing approaches on unaligned multimodal sequences but also has strong performance on aligned multimodal sequences.
\end{abstract}

\section{Introduction}

Multimodal sentiment analysis has recently become a widely researched topic in natural language and multimodal machine learning communities ~\cite{IntelligentSystems-2016-Zadeh-Zellers-multimodal,InformationFusion2017-MultimodalFusion,Language2008-IEMOCAP,NIPS2019_MultilinearFusion,ICLR-2019-factorized-multimodal-representations,ACL-2018-bagher-zadeh-multimodal,ACL2017-ContextSentiment,emnlp-2018-liang-multimodal,AAI2019-WordShift,AAAI2019-Translation-Modal}. Intrinsically, the way of people expressing their opinions and sentiments involves multiple modalities including the language (words), visual (facial expressions and head gestures), and acoustic
(paralinguistic), which are in the form of asynchronous coordinated sequences. In particular, Tsai et al.~\cite{ACL2019-MultmodalTransformer} have clearly illustrated there is an ``unaligned'' nature of the multimodal sentiment analysis task.
The receiving frequencies of receptors usually vary in audio and
vision streams, and hence it is difficult to obtain optimal
alignment between them without manual data preprocessing. For example, a frowning face may relate to a pessimistically word spoken in the past or coming clips. The heterogeneities and unaligned nature across modalities often increase the difficulty of analyzing multimodal sequences. 

\begin{figure*}[!h]
	\centering
	\subfigure[MCMulT architecture.]
	{
		\label{fig:framework-a}
		\begin{minipage}[t]{0.31\linewidth}
			\centering
			\includegraphics[width=1.1\textwidth]{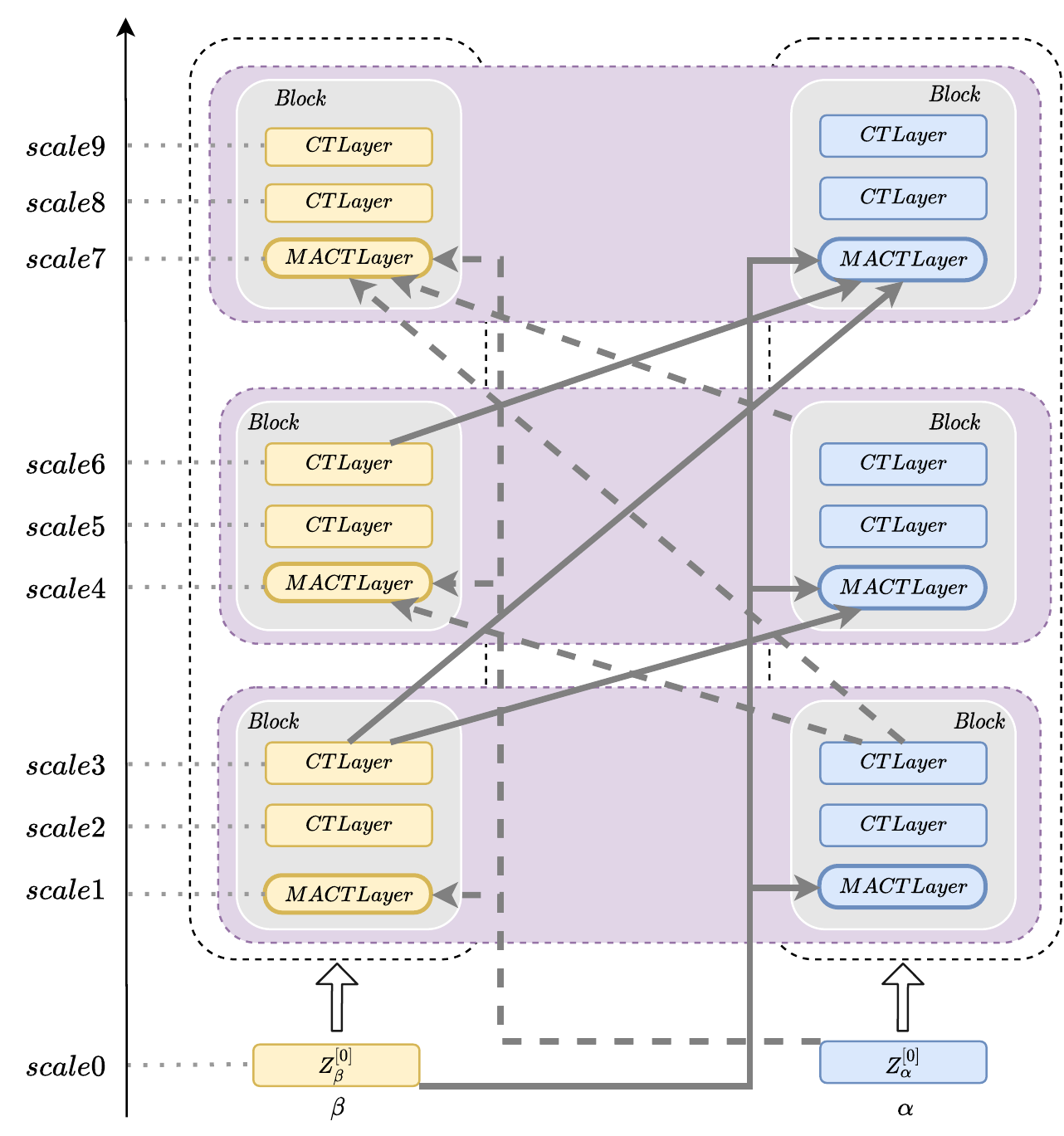}
		\end{minipage}
	}\subfigure[Details of block mechanism.]
	{
		\label{fig:framework-b}	
		\begin{minipage}[t]{0.31\linewidth}
			\centering
			\includegraphics[width=0.7\textwidth]{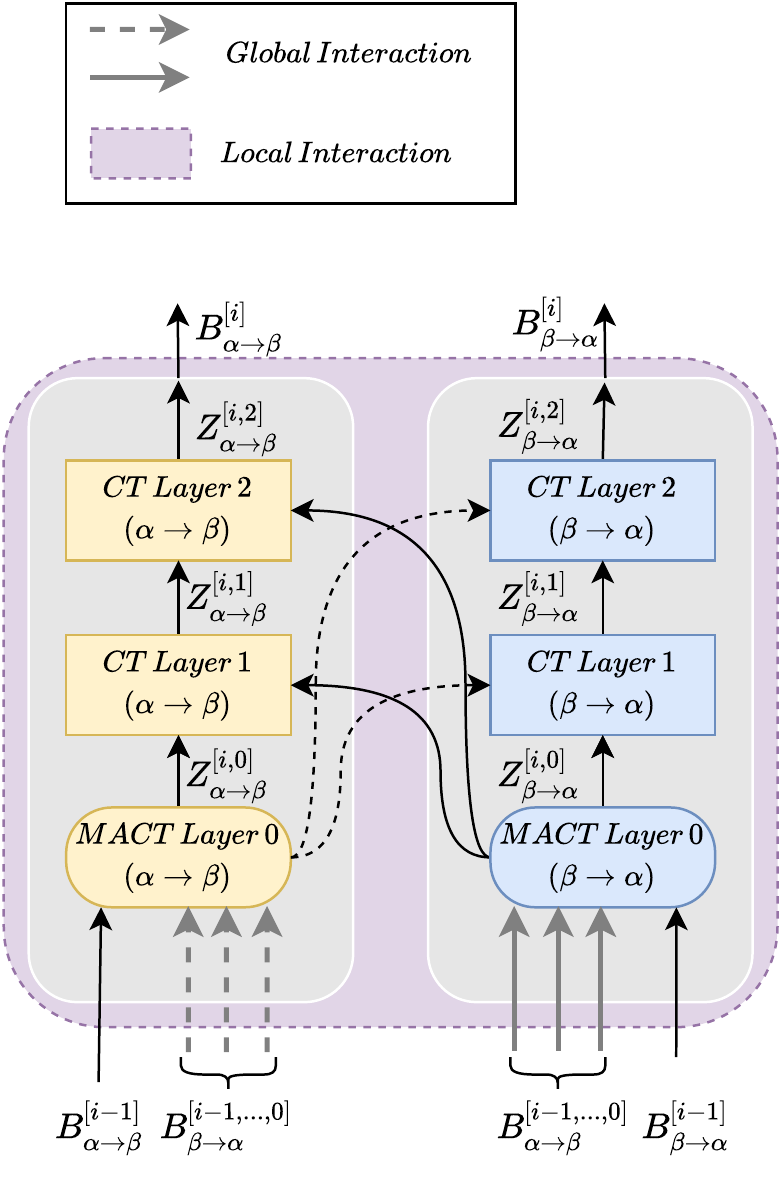}
		\end{minipage}
	}\subfigure[Vanilla MulT~\cite{ACL2019-MultmodalTransformer}.]
	{	
		\label{fig:framework-c}
		\begin{minipage}[t]{0.31\linewidth}
			\centering
			\includegraphics[width=1.1\textwidth]{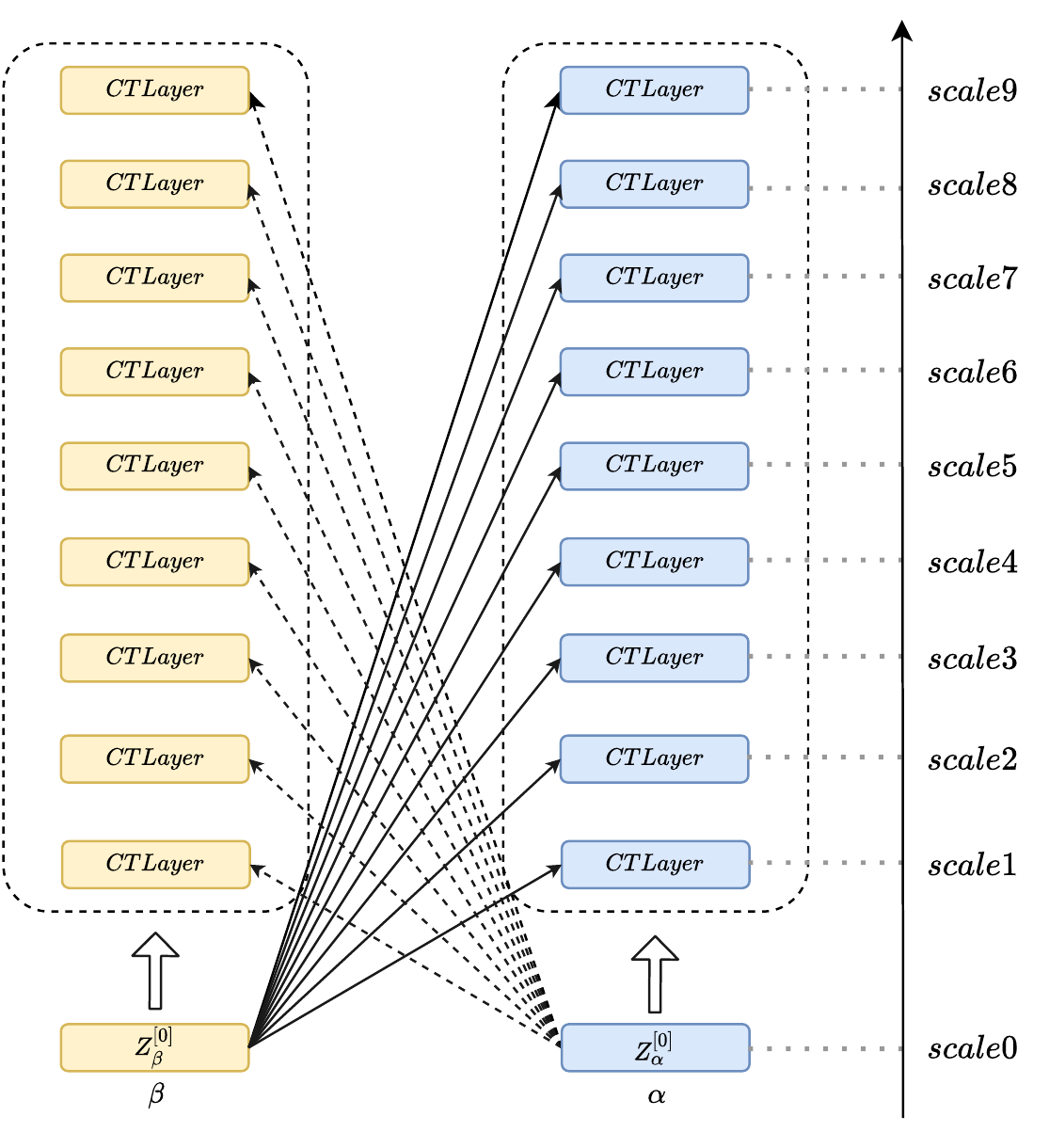}
		\end{minipage}
	}
	\caption{Illustration of our Multi-scale Cooperative Multimodal Transformer (MCMulT) and the vanilla Multimodal Transformer (MulT) architecture. For one given target modality, the directional crossmodal interaction is implemented via a crossmodal transformer layer (CT layer) with the representation from its source modality. MACT extends CT layer with attention on multi-scale representations of the source modality. Bidirectional interactions are performed in a cooperative manner. One gray region of Figure~\ref{fig:framework-a} denotes the multi-scale crossmodal Transformer block (MCTB) that contains one MACT and multiple CT Layers. (Best viewed in color and larger version)}	
	\label{fig:framework}
\end{figure*}

To tackle these challenges, several papers~\cite{NIPS2019_MultilinearFusion,ICLR-2019-factorized-multimodal-representations,emnlp-2018-liang-multimodal,ACL2017-ContextSentiment,ACL2020Challenge-JointTransformer,ACL2019-MultmodalTransformer} perform multimodal fusion in various manners. Initial works~\cite{emnlp-2018-liang-multimodal,ACL2017-ContextSentiment,NIPS2019_MultilinearFusion,ACL2020Challenge-JointTransformer,ACL2020-Multilogue-Net,ACL2020-MultimodalRouting} in unaligned multimodal sentiment analysis use a common way to manually force word-aligning before training and testing. Specifically, the visual and acoustic features are first manually aligned to
the resolution of textual words. It is impractical to manually align different streams in the real-word scenario. Besides, as the analysis in~\cite{ACL2019-MultmodalTransformer}, the word-aligned approaches not only depend on feature engineering that involves domain knowledge, but are also inclined to overlook long-range crossmodal contingencies of the original modalities. To make the alignment and fusion over multiple modalities be more feasible in practice, Tsai et al.~\cite{ACL2019-MultmodalTransformer} proposes a crossmodal attention module by extending the standard Transformer network~\cite{NIPS2017-Transformer} to learn representations directly from unaligned multimodal sequences.

In this paper, we propose a multi-scale cooperative multimodal transformer (MCMulT) architecture to improve the quality of representations learned from unaligned multimodal sequences. 
The intuition is that when the multimodal sequences is unaligned, the information from different modalities can be very different. To improve the robustness of a transformer, during the crossmodal interaction, as much as the information from other modalities should be kept. Therefore, in our method, mutli-scale representations that contain different levels of semantic information are used for crossmodal interaction instead of a single-scale representation.  
Furthermore, we designed a computationally-efficient way to let the representations from different scales and different modalities communicate with each other. As a result, different levels of information from different modalities are captured and synchronized by the transformers, which improves the performance of the transformers and the quality of the learned representations.
As shown in Figure~\ref{fig:framework}, our method let all modalities progressively learn their individual hierarchical representations by exploiting directional pairwise crossmodal interactions with the multi-scale representations on hand. Each crossmodal interaction can be regarded as one feature enhancement operator (MACT Layer in Figure~\ref{fig:framework-a}) that attends the long-term spatial dependencies from all the preceding transformer layers of the source modality. This not only shortens the path of crossmodal error propagation, but also helps to prevent the gradually evolving interactions from being forgotten or diluted. Consequently, the enhanced layer of target modality is also used as input to boost all subsequent layers of its source modalities. Following this way, a pair of modalities respectively build their hierarchical representations in a cooperative manner. Furthermore, to improve effectiveness and computational efﬁciency, instead of forcing fully dense interactions, we introduce a block-scale mechanism that is a combination of a windowed local-context and global attention interaction through ablations and controlled trials. This reduces complexity of the network while bringing higher performance.

We conduct extensive experiments on unaligned and aligned multimodal sentiment analysis, covering three benchmarks: CMU-MOSI~\cite{IntelligentSystems-2016-Zadeh-Zellers-multimodal}, CMU-MOSEI~\cite{ACL-2018-bagher-zadeh-multimodal} and IEMOCAP~\cite{Language2008-IEMOCAP}. Our experiments show that our MCMulT outperforms previous work on both the common word-aligned setting and the more challenging unaligned scenario. In addition, empirical qualitative analysis further proves that the block-scale mechanism brings improvements on performance via increasing the depth of crossmodal networks compared to other variants of MCMulT.

\section{Related Works}
\textbf{Sentiment Analysis.} Early work on sentiment analysis or emotion recognition focused primarily on one modality, i.e., text~\cite{ACL2002-SemanticOrientation,ACL2004-SentimentalEducation,EMNLP2013-SentimentTreebank}, vision ~\cite{Culture1974_FacialExpressionsn} and
audio~\cite{ICASSP2012-AudioEmotion}. Probably the most challenging
task in multimodal sentiment analysis is learning a good representation of multiple modalities. More researchers have committed to integrating the multimodal information effectively. To date, there are mainly two types
of fusion strategies: early fusion and late fusion. Methods in the early fusion category concatenate multimodal data at the input level~\cite{IntelligentSystems-2016-Zadeh-Zellers-multimodal,ICME2017-MultimodalSentiment}. While early fusion methods outperform unimodal models, they cannot comprehensively cover the modality-specific interactions
and tend to overfit. The late fusion methods integrate different modalities after input stage, and then exploit both modality-specific and crossmodal interactions~\cite{EMNLP2017-TensorFusion,ICLR-2019-factorized-multimodal-representations,HML2018-Seq2SeqSentiment,ACL2020-ContextRNN,ACL2020Challenge-JointTransformer,ACL2020-Multilogue-Net,ACL2020-MultimodalRouting}. Currently, several competitive results are achieved by augmenting this class of models with attention or memory mechanism~\cite{emnlp-2018-liang-multimodal,AAI2019-WordShift,ACL2020Challenge-JointTransformer}. Our work follows the attention mechanisms to model intra-modal or inter-modal interactions from multimodal sequences.

\textbf{Transformer Model.}
Transformers were designed by Vaswani et al.~\cite{NIPS2017-Transformer} as a novel attention-based building block for modeling sequential data. Recently, Transformer models have been successfully applied to machine learning community including natural language processing, speech processing and computer vision~\cite{ACL2019-BERT,ICML2018-ImageTransformer,carion2020end,ICASSP2018-SpeechTransformer}. Inspired by these Transformer scaling successes, several recent works have also introduced transformer-based alignment or fusion to model relations between different modalities.
For example, the popular BERT architecture~\cite{ACL2019-BERT} is extended to learn joint visual-linguistic representations in a self-supervision pretraining framework~\cite{NIPS2019-ViLBERT,EMNLP-IJCNLP2019-LXMERT,ECCV2020-UNITER,ACL2020-Pretrain}. For VQA task, Hu et al.~\cite{CVPR2010-TransformersVQA} project all entities from different modalities (query words, objects in the image and OCR results
of the image) into a common semantic embedding space and apply Transformer to collect relational representations for each entity. These multimodal transformer approaches mostly input all modalities either independently or jointly to the vanilla Transformer module without
explicitly exploiting both multi-scale and cooperative mechanism in a holistic perspective especially for the unalign nature in the multimodal squences. Delbrouck et al.~\cite{ACL2020Challenge-JointTransformer} describes a
Transformer-based joint-encoding (TBJE) for sentiment analysis task on the manually aligned multimodal sequences. To tackle the unaligned multimodal sentiment analysis task, Tsai et al.~\cite{ACL2019-MultmodalTransformer} propose a directional pairwise crossmodal attention in the multimodal transformer (vanilla MulT\footnote{We use MulT to indicate the multimodal Transformer ~\cite{ACL2019-MultmodalTransformer}.}) model to exploit interactions between multimodal sequences across different time steps. Our work has the same cornerstone with MulT. However, there are two main differences between MultT (or TBJE) and our MCMulT. Firstly, MulT models crossmodal interactions only use the single scale (the low-level) features from its source modality. In our MCMulT,
multi-scale features are exploited to model each crossmodal interaction, and the sophisticated crossmodal interactions also facilitate pairwise modalities' ablity to learn progressive multi-scale features in a cooperative manner.
Secondly, we design a block-scale mechanism in the MCMulT which allows more transformer layers to improve the recognition rate. The block-scale mechanism is not conducted in MulT.

\section{Proposed Method}

\begin{figure*}[t]
	\centering
	\subfigure[Multimodal classification with MCMulT.]
	{
		\label{fig:FusionNet}
		\begin{minipage}[t]{0.31\linewidth}
			\centering
			\includegraphics[width=1\linewidth,height=0.18\pdfpageheight]{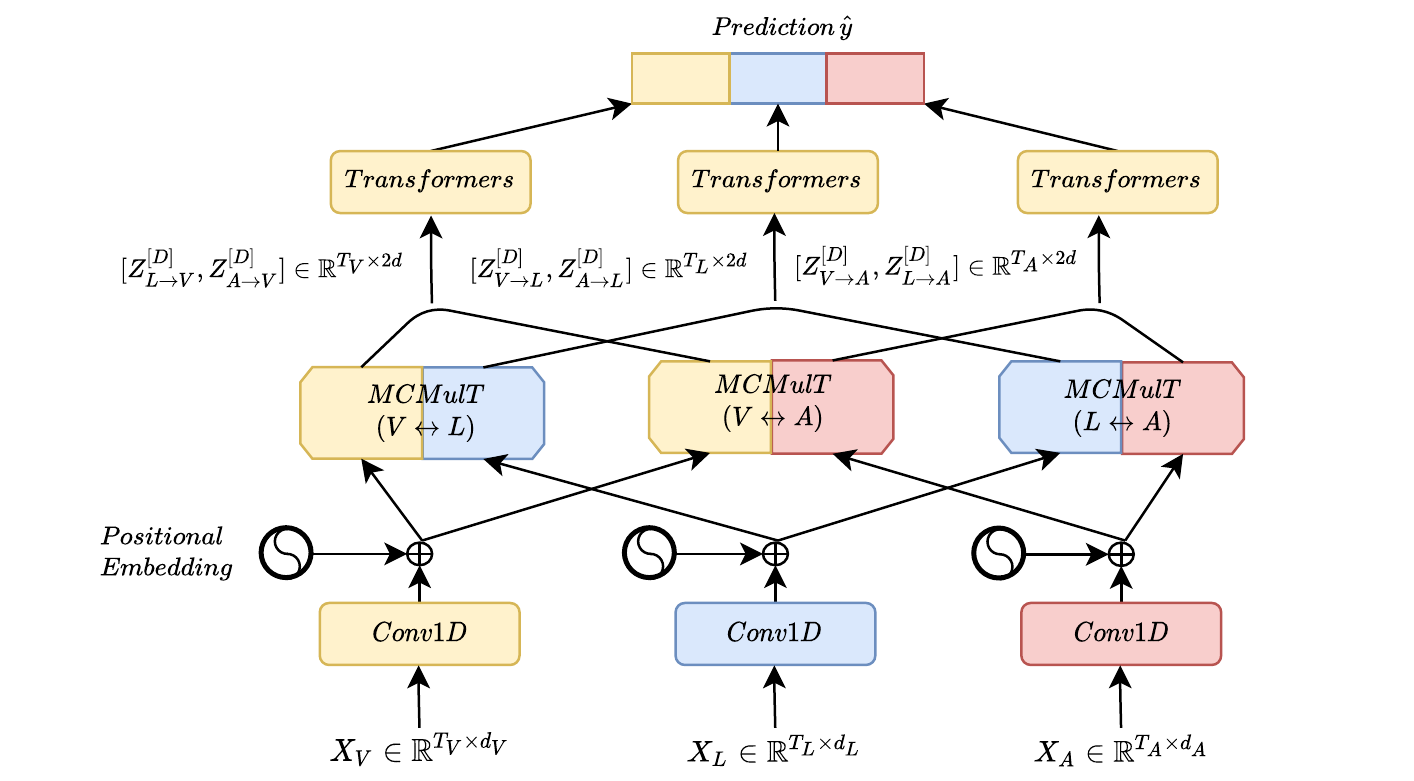}
		\end{minipage}
	}\subfigure[MulT~\cite{ACL2019-MultmodalTransformer}.]
	{
		\label{fig:MACT}
		\begin{minipage}[t]{0.31\linewidth}
			\centering
			\includegraphics[width=1.0\textwidth,height=0.18\pdfpageheight]{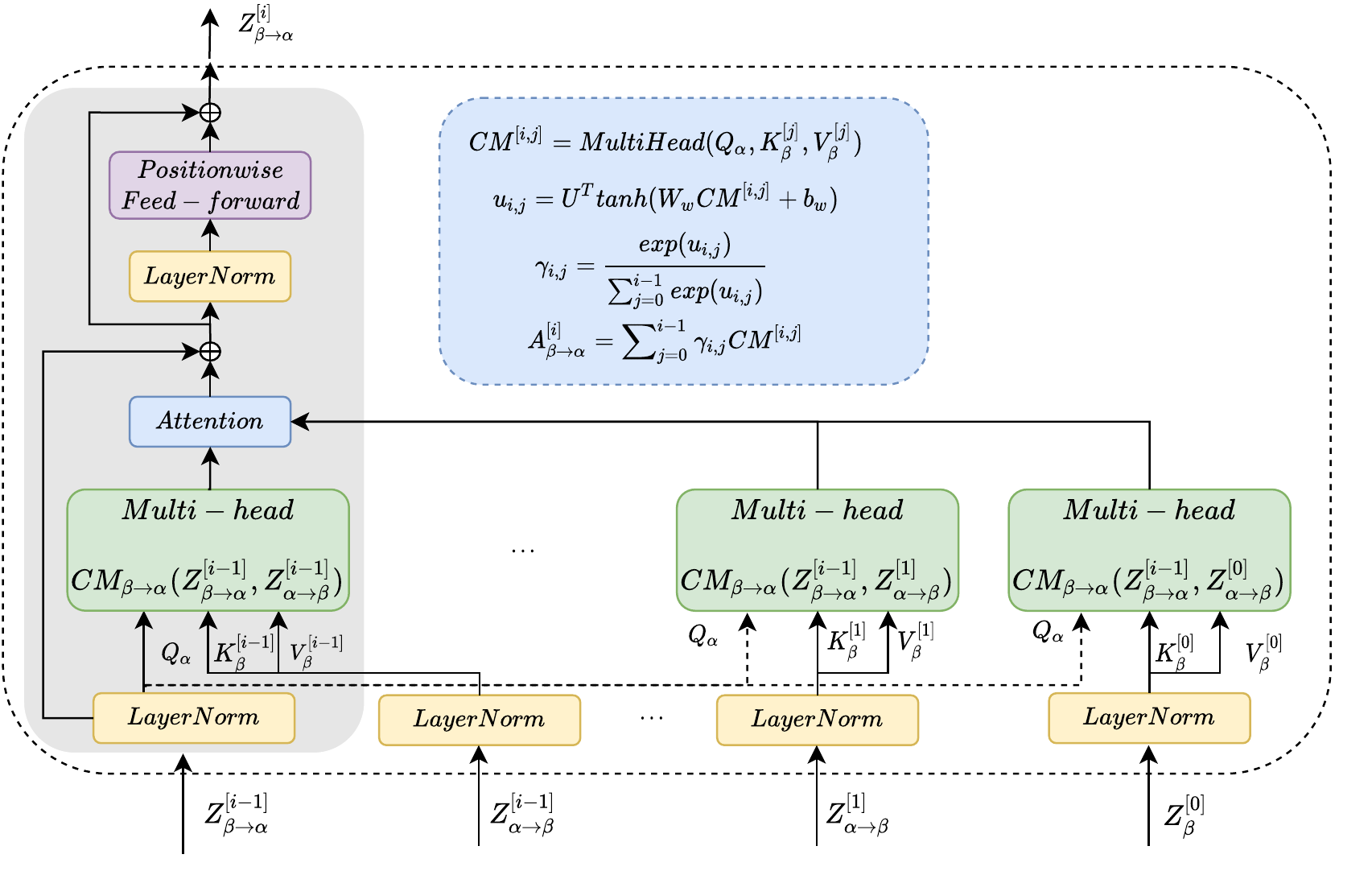}
		\end{minipage}
	}\subfigure[CT unit~\cite{ACL2019-MultmodalTransformer}.]
	{	
		\label{fig:CT}
		\begin{minipage}[t]{0.31\linewidth}
			\centering
			\includegraphics[width=1.\textwidth,height=0.18\pdfpageheight]{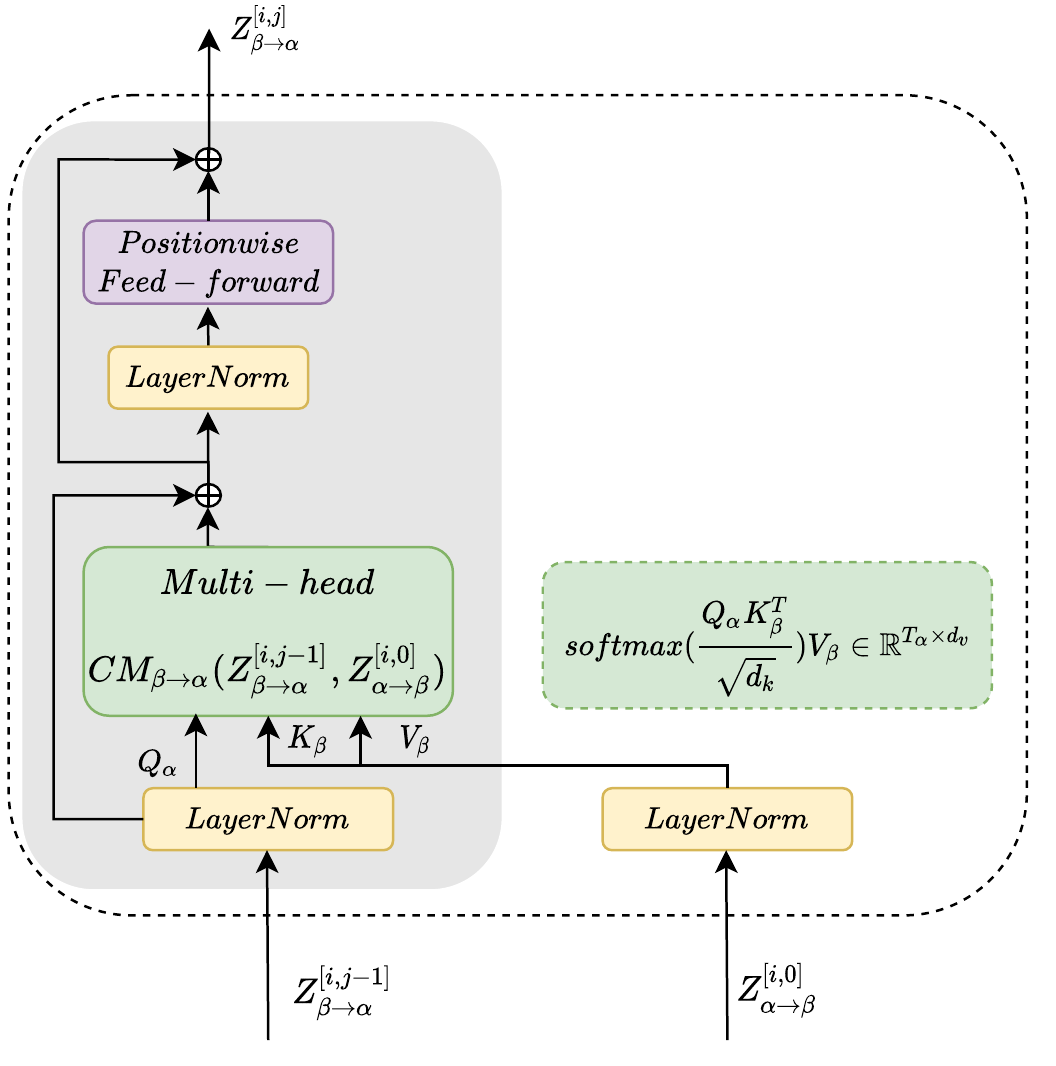}
		\end{minipage}
	}

\end{figure*}

In this section, we describe our proposed MCMulT architecture for unaligned multimodal sequences sentiment analysis task, as shown in Figure~\ref{fig:FusionNet}. Similarly with MulT ~\cite{ACL2019-MultmodalTransformer}, the representations are built from multiple pairwise modalities, e.g. text-vision, text-audio, vision-audio, and then merged at the high level to predict the sentiment. The network contains three modules: low-level feature, MCMulT and prediction module. MCMulT is the core component of the network which explicitly exploits both multi-scale and cooperative mechanisms for multimodal fearture learning in a holistic perspective. Specifically, each crossmodal transformer layer attends all the preceding transformer layers (not only the low-level features) of its source modality. In this way, MCMulT learns progressive multi-scale features with the built feature hierarchies at hand and coordinates them through attentive crossmodal interactions. As a result, different levels of information from different modalities are captured, and the quality of the learned representations are improved. We first describe low-level feature module in Section3.1. In Section 3.2, the proposed core module MCMulT is presented in detail, and some variants of MCMulT are also discussed. Finally, the prediction module is briefly illustrated in Section 3.3.

\subsection{Low-level Features}
Temporal convolution and positional embedding are applied to extract the low-level features from the original three modalities as~\cite{ACL2019-MultmodalTransformer} including text (L), vision (V) and audio (A). Let $X_{\{L,V,A\}} \in R^{T_{\{L,V,A\}} \times d_{\{L,V,A\}}}$ represent the raw data from three-modality sequences where $T$ and $d$ respectively indicate the sequence length and the data dimension. 

Temporal convolution is expected to exploit the local structure information of the sequence and project the features of different modalities to the same dimension $d$, which is performed with a one-dimensional convolution layer:

\begin{equation}\label{equation:1d-convolution}
\resizebox{.9\hsize}{!}{
	$\hat{X}_{\{L,V,A\}}=Conv1D(X_{\{L,V,A\}},k_{\{L,V,A\}})\in R^{T_{\{L,V,A\}} \times d},$
	}
\end{equation}
where $k_{\{L,V,A\}}$ denotes the convolution kernel, $d$ is the common dimension.

Following Vaswani et al.~\cite{NIPS2017-Transformer}, positional embedding is augmented to the output of temporal convolution as Equation~\ref{equation:position-embedding} , which helps carrying temporal information from each modality sequence. 
For more details of the positional embedding, please refer to ~\cite{NIPS2017-Transformer,ACL2019-MultmodalTransformer}. $Z^{[0]}_{\{L,V,A\}}$ denotes the low-level features that will be fed to our MCMulT module:

	\begin{equation}\label{equation:position-embedding}
		Z^{[0]}_{\{L,V,A\}}=\hat{X}_{\{L,V,A\}}+PE(T_{\{L,V,A\}}),
	\end{equation}
where $PE(\cdot)$ indicates the positional embedding.

\subsection{MCMulT}\label{sec:MCMulT}

The core idea of MCMulT is that directional crossmodal interactions are performed in the multi-scale and cooperative mechanisms as shown in Figure~\ref{fig:framework-a}, where both solid and dotted arrows indicate directional crossmodal interactions between a pair of modalities.

Multi-scale mechanism indicates that a target modality builds its multi-scale features via integrating directional crossmodal interactions. Each integration is accomplished through aggregating multi-scale features of its source modality, as illutrated in Figure~\ref{fig:framework-a}.

Cooperative mechanism indicates that our MCMulT allows the source and target modality iteratively boost each other during building their multi-scale features, as shown in Figure~\ref{fig:framework-a}, where the information flow is bi-directional according to the black solid and dotted arrows. Specifically, as Figure~\ref{fig:framework-a} illustrates, after the scale-1 representation of modality $\alpha$ enhances the scale-2 representation of modality $\beta$, the enhanced scale-2 representation of modality $\beta$ also is used to boost the subsequent representations (scale-3 and scale-4) of modality $\alpha$
following the reciprocity principle.

\subsubsection{MCTB}
The MCMulT network is divided into multiple densely connected multi-scale crosmodal Transformer blocks (MCTBs) which contain two types of crossmodal units: multi-scale attentive crossmodal transformer (MACT) and crossmodal transformer (CT). The core unit is MACT which contains three sub-network layers: muti-scale multi-head crossmodal layer, multi-scale attentive layer and position-wise feed-forward layer, as shown in Figure~\ref{fig:MACT}. CT only attends single scale representation of the source modality, and can be regarded as a simple version of MACT.


\begin{figure*}[!h]
	\centering
	\subfigure[MCMulT.]
	{
		\label{fig:variants-a}
		\begin{minipage}[t]{0.25\linewidth}
			\centering
			\includegraphics[width=0.9\textwidth]{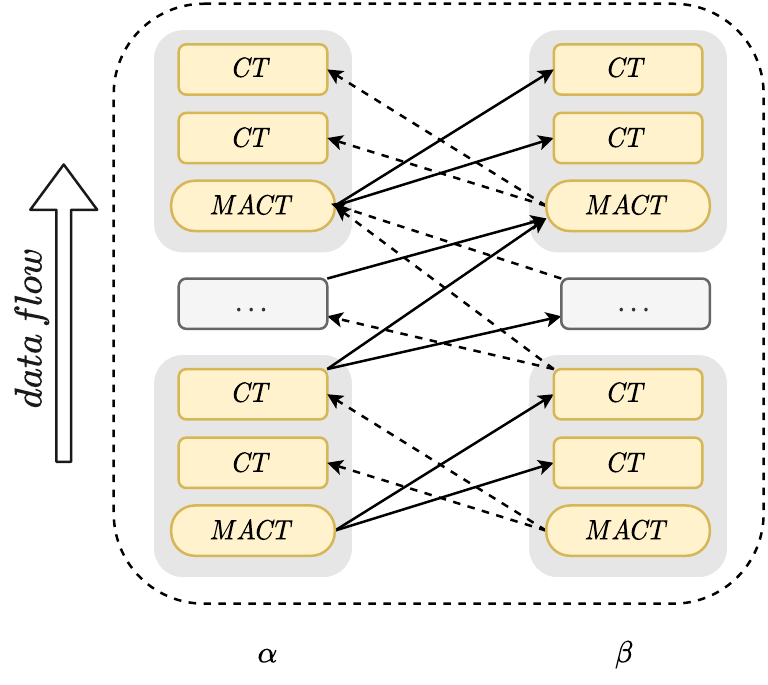}
		\end{minipage}
	}\subfigure[MCMulT-Dense.]
	{
		\label{fig:variants-b}
		\begin{minipage}[t]{0.25\linewidth}
			\centering
			\includegraphics[width=0.9\textwidth]{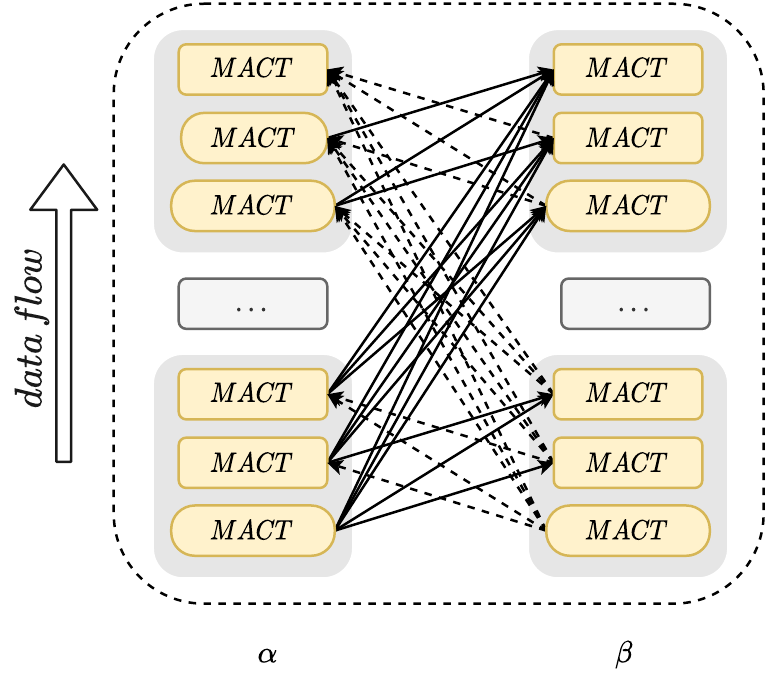}
		\end{minipage}
	}\subfigure[MCMulT-LocalDense.]
	{
		\label{fig:variants-c}	
		\begin{minipage}[t]{0.25\linewidth}
			\centering
			\includegraphics[width=0.9\textwidth]{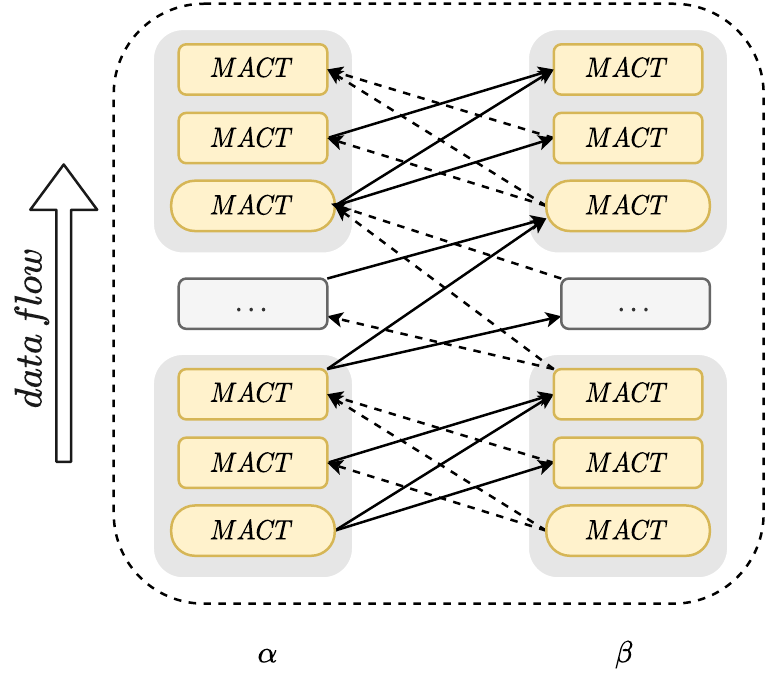}
		\end{minipage}
	}\subfigure[MCMulT-Global.]
	{	
		\label{fig:variants-d}
		\begin{minipage}[t]{0.25\linewidth}
			\centering
			\includegraphics[width=0.9\textwidth]{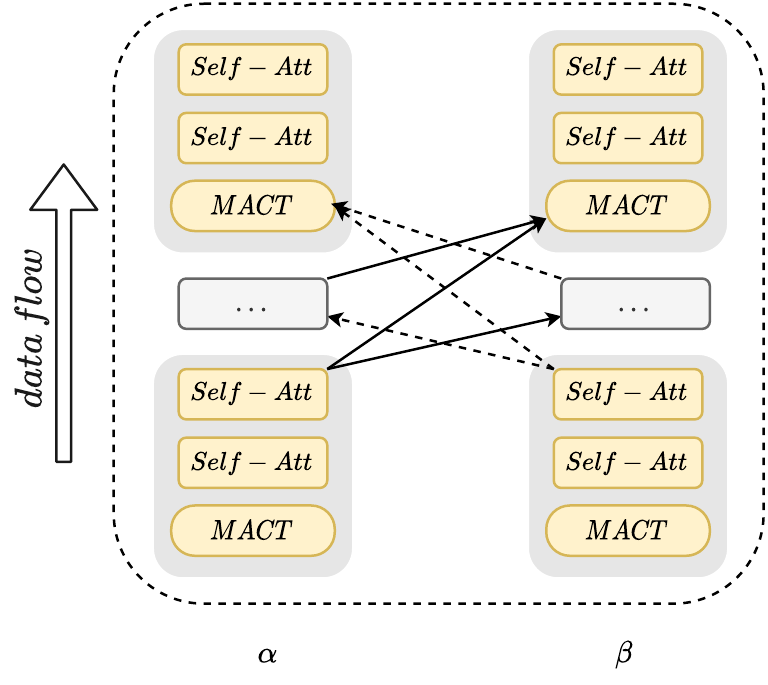}
		\end{minipage}
	}
	\caption{Illustration of MCMulT variants.}	
	\label{fig:mcmult_variants}
\end{figure*}

\textbf{Multi-scale Crossmodal Layer.} Let $Z_{\alpha} \in R^{T_{\alpha} \times d_{\alpha}}$ and $Z_{\beta} \in R^{T_{\beta} \times d_{\beta}}$ represent the features (any scale or level) from two modality sequences respectively. As shown in the green region of Figure~\ref{fig:MACT}, one multi-scale crossmodal layer aggregates multiple directional pairwise crossmodal interactions between the target modality $\alpha$ and the source modality $\beta$ via multiple multi-head layers in parallel. Each of crossmodal interactions is detailed in Equation~\ref{equation:multi-head} as the vanilla transformer~\cite{NIPS2017-Transformer}.

\begin{equation}\label{equation:multi-head}
\resizebox{.9\hsize}{!}{	
		$CM_{\beta\rightarrow\alpha}(Z_\alpha,Z_\beta)=softmax\left(\frac{Z_\alpha W_{Q_\alpha} W_{K_\beta}^T Z_\beta^T}{\sqrt{d_k} }\right)Z_\beta W_{V_\beta},$
}
\end{equation}
where $W_{Q_\alpha}$, $W_{K_\beta}$ and $W_{V_\beta}$ are weight parameters. When the modality $\alpha$ builds its $i$-th level crossmodal interactions from the modality $\beta$, $H^{[i]}$ denotes the set of multi-scale crossmodal interactions between the modality $\alpha$ and $\beta$ as

	\begin{equation}\label{equation:multi-head-set}
		\begin{aligned}
			&H^{[i]}=\left\{CM_{\beta\rightarrow\alpha}(Z_{\beta\rightarrow\alpha}^{[i-1]},Z_{\alpha\rightarrow\beta}^{[j]})\right\},j=0,...,i-1 \\
			&Z_{\alpha\rightarrow\beta}^{[0]}=Z_{\beta}^{[0]},
		\end{aligned}
	\end{equation}
where $Z_{\beta}^{[0]}$ is the low-level feature of the modality $\beta$ as Equation \ref{equation:position-embedding}.

\textbf{Multi-scale Attention.} To ensure the relevant information of each crossmodal interaction of $H^{[i]}$, $H^{[i]}$ is aggregated with multi-scale attention to generate an enhanced feature $A_{\alpha\rightarrow\beta}^{[i]}$ with a self-attention component as shown in the blue part of Figure~\ref{fig:MACT}.

\textbf{Positionwise Feed-Forward Layer}. Following the vanilla Transformer~\cite{NIPS2017-Transformer}, we input the attentive fusion feature to the positional feed-forward layer in Euqation~\ref{equation:PPF} corresponding to the purple part of Figure~\ref{fig:MACT}.

	\begin{equation}\label{equation:PPF}
		\begin{aligned}
			&P^{[i]}_{\beta\rightarrow\alpha}=f_\theta(LN(A_{\alpha\rightarrow\beta}^{[i]}+LN(Z_{\beta\rightarrow\alpha}^{[i-1]}))) \\
			&Z_{\beta\rightarrow\alpha}^{[i]}=\{A_{\alpha\rightarrow\beta}^{[i]}+LN(Z_{\beta\rightarrow\alpha}^{[i-1]})\}+P^{[i]}_{\beta\rightarrow\alpha}
		\end{aligned}
	\end{equation}

\subsubsection{Block Interaction Mechanism}
Intuitively, it is possible to only use MACT layer to build multi-scale crossmodal interactions in the MCMulT module. However, this mechanism with dense multi-scale interaction also brings more parameters than the single-scale crossmodal interaction.
Especially, this mechanism will make the network complexity grow linearly with the network depth, and the training will become much harder. To address the above limitation, we dive into the network connection structure and replace the dense connections with sparse connections, which gives birth to the block mechanism. There are two kinds of crossmodal interaction in a MCTB, i.e., global interactions and local interactions as shown in Figure~\ref{fig:framework-a} and Figure~\ref{fig:framework-b}.

To attend multi-scale dependencies, global interactions are built densely between the target and source modalities in the block scale. Each block of MCMulT involves multiple crossmodal Transformer layers. The global interacions of MCMulT are performed  with  MACTs and the input of MCTB is from the output of multiple MCTBs of the source modality. To preserve a windowed local context, local interactions are built between the MCTB of target modality and the same-scale MCTB of its source modality. Local interactions of MCMulT are performed with a CT as shown in the Figure~\ref{fig:CT}. The input of CT contains the previous layer output of the target modal and the first layer output of the same level MCTB from its source modality. It is noted that the local interaction only uses the single scale representation from source modality. This block mechanism makes the number of parameters decrease.

\subsubsection{Variants of MCMulT} Block mechanism is only one manner to explore multi-scale interactions betwwen the source and target modality in MCMulT. If different network connection structures are designed to take the place of block structure, MCMulT can be changed into three variants: MCMulT-Dense, MCMulT-LocalDense and MCMulT-Global as shown in Figure~\ref{fig:mcmult_variants}. MCMulT with the dense connections from all the preceding layers is denoted as MCMulT-Dense. MCMulT-LocalDense models local interactions in a dense manner with preserving global interactions of MCTBs. MCMulT-Global is obtained through cutting out local crossmodal interactions. It is noted that MulT may be regarded as an extremly simple version of MCMulT-Dense. If MCMulT-Dense only uses the scale-0 of source modality to build crossmodal interactions, MCMulT-Dense will degrade to MulT. The descending order of complexity of MCMulT variants is as MCMulT-Dense, MCMulT-LocalDense, MCMulT, MulT, MCMulT-Global. The performances of these variants of MCMulT will be discussed in the experiments.

\subsection{Prediction}
As the final step, we fuse the features of all modalities through concatenating the outputs from MCMulT architecture that share the same target modality to yield $\tiny{\left[Z_{V\rightarrow L}^{[D]},Z_{A\rightarrow L}^{[D]}\right]}$, $\tiny\left[Z_{L\rightarrow V}^{[D]},Z_{A\rightarrow V}^{[D]}\right]$, $\left[Z_{L\rightarrow A}^{[D]},Z_{V\rightarrow A}^{[D]}\right]$. Each item  is then passed through a different Transformer to collect temporal information. Eventually, the outputs of transformers are extracted to pass through a full-connected layer to predict sentiment.

\section{Experiments}
In this section, we conduct experimental evaluation of MCMulT on three datasets (CMU-MOSI, CMU-MOSEI, IEMOCAP), which are often used as benchmarks for multimodal sentiment analysis tasks. Because of the limited space, we leave out details of implementations in Appendix~A and a visualization of crossmodal attention maps in Appendix~B. Our experiments are mainly divided into two parts. Firstly, our MCMulT is compared with the existing competitive approaches in both word-unaligned and word-aligned settings. Secondly, the variants of our MCMulT and related hyperparameters are evaluated.

\subsection{Datasets and Evaluation Metrics}
Each task is conducted in both word-aligned and unaligned settings. For both settings, the multimodal raw features are extracted from the textual (Glove word embedding~\cite{EMNLP2014-Glove}), visual (Facet~\cite{iMotions}) and acoustic (COVAREP~\cite{ICASSP2014-COVAREP}) modalities. For the unaligned setting, we use the raw original audio and visual features as extracted, without any word-segmented alignment or manual subsampling. For the word-aligned case, all modals are aligned by P2FA~\cite{JASA2008-P2FA} as did in ~\cite{ICLR-2019-factorized-multimodal-representations,AAAI2019-Translation-Modal,AAI2019-WordShift,ACL2019-MultmodalTransformer}. A more detailed introduction can be found in MulT~\cite{ACL2019-MultmodalTransformer}.

\textbf{CMU-MOSI.} CMU-MOSI~\cite{IntelligentSystems-2016-Zadeh-Zellers-multimodal} dataset consists of 2,199 opinion video clips from YouTube movie reviews spoken in English. Each clip is annotated with sentiment in the range [-3,3] from high negative to high positive. Sentiment analysis task on CMU-MOSI is considered as a 7 class sentiment classification problem. In particular, 1,284, 229 and 686 clips are used for training, valid and test, respectively.

\textbf{CMU-MOSEI.} CMU-MOSEI~\cite{ACL-2018-bagher-zadeh-multimodal} consists of 23,454 movie review video clips from YouTube spoken in Spanish, each of which is annotated with a sentiment score between -3 (strongly negative) and +3 (strongly positive) to indicate emotional preferences. 16,326, 1,871 and 4659 utterances are used for training, valid and test respectively.

\textbf{IEMOCAP.} IEMOCAP~\cite{Language2008-IEMOCAP} consists of two-way conversations between 10 speakers, which are divided into utterances. As suggested by Wang et al. ~\cite{AAI2019-WordShift}, the utterances are tagged with the labels happy, sad, angry and neutral. 2,717, 798 and 938 utterances are used for training, valid and test respectively. 

\textbf{Evaluation Metrics.} For MOSEI and MOSI datasets, the same metrics were used to evaluate the performance of the model: 7-class accuracy (i.e. emotion score classification in Acc7: [-3, +3]), binary accuracy (i.e. Acc2: positive/negative emotion), F1 score, mean absolute error (MAE) of score and correlation of model prediction with human. In IEMOCAP dataset, binary accuracy and F1 values are used to evaluate the performance of the model.

\begin{table}[!t]
	\scriptsize
	\centering
	\setlength{\tabcolsep}{0.75mm}{
		\begin{tabular}	{|c|c|c|c|c|c|} 
			\hline
			Metric & $\text{Acc}^h_7$ & $\text{Acc}^h_2$ & $\text{F1}^h$ & $\text{MAE}^l$ & $\text{Corr}^h$\\ \hline
			\hline
			\multicolumn{6}{|c|}{CMU-MOSI-Aligned}\\ \hline
			\hline
			EF-LSTM & 33.7 & 75.3 & 75.2 & 1.023 & 0.608\\ \hline
			LF-LSTM & 35.3 & 76.8 & 76.7 & 1.015 & 0.625\\ \hline
			RMFN~\cite{emnlp-2018-liang-multimodal} & 38.3 & 78.4 & 78.0 & 0.922 & 0.681\\ \hline
			MFM~\cite{ICLR-2019-factorized-multimodal-representations} & 36.2 & 78.1 & 78.1 & 0.951 & 0.662\\ \hline
			\hline
			RAVEN~\cite{AAI2019-WordShift} & 33.2 & 78.0 & 76.6 & 0.915 & 0.691\\ \hline
			MCTN~\cite{AAAI2019-Translation-Modal} & 35.6 & 79.3 & 79.1 & 0.909 & 0.676\\ \hline
			Mu-Net~\cite{ACL2020-Multilogue-Net} & - & 81.2 & 80.1 & - & -\\ \hline
			MulT~\cite{ACL2019-MultmodalTransformer} & 40.0 & 83.0 & 82.8 & 0.871 & 0.698\\ \hline
			\hline
			MCMulT (ours) & \textbf{40.7} & \textbf{83.9} & \textbf{83.2} & \textbf{0.866} & \textbf{0.701}\\ \hline
			\hline
			\multicolumn{6}{|c|}{CMU-MOSI-Unaligned}\\ \hline
			CTC~\cite{ICML2016-CTC}+EF-LSTM& 31.0 & 73.6 & 74.5 & 1.078 & 0.542\\ \hline
			LF-LSTM & 33.7 & 77.6 & 77.8 & 0.988 & 0.624\\ \hline
			CTC+RAVEN~\cite{AAI2019-WordShift} & 31.7 & 72.7 & 73.1 & 1.076 & 0.544\\ \hline
			CTC+MCTN~\cite{AAAI2019-Translation-Modal} & 32.7 & 75.9 & 76.4 & 0.991 & 0.613\\ \hline
			MulT~\cite{ACL2019-MultmodalTransformer} & 39.1 & 81.1 & 81.0 & 0.889 & 0.686\\ \hline\hline
			MCMulT (ours) & \textbf{40.3} & \textbf{82.2} & \textbf{82.3} & \textbf{0.885} & \textbf{0.691}\\ \hline
		\end{tabular}
	}
	\caption{Results on CMU-MOSI with aligned and unaligned multimodal sequences. $h$ means higher is better and $l$ means lower is better. EF stands for early fusion, and LF stands for late fusion.}
	\label{tab:result-CMU-MOSI}
\end{table}	
\begin{table}[!t]
	\scriptsize
	\centering
	\setlength{\tabcolsep}{0.75mm}{
	\begin{tabular}	{|c|c|c|c|c|c|} 
			\hline
			Metric & $\text{Acc}^h_7$ & $\text{Acc}^h_2$ & $\text{F1}^h$ & $\text{MAE}^l$ & $\text{Corr}^h$\\ \hline
			\hline
			\multicolumn{6}{|c|}{CMU-MOSEI-Aligned}\\ \hline
			\hline
			EF-LSTM & 47.4 & 78.2 & 77.9 & 0.642 & 0.616\\ \hline
			LF-LSTM & 48.8 & 80.6 & 80.6 & 0.619 & 0.659\\ \hline
			Graph-MFN~\cite{ACL-2018-bagher-zadeh-multimodal} & 45.0 & 76.9 & 77.0 & 0.71 & 0.54\\ \hline
			RAVEN~\cite{AAI2019-WordShift} & 50.0 & 79.1 & 79.5 & 0.614 & 0.662\\ \hline
			MCTN~\cite{AAAI2019-Translation-Modal} & 49.6 & 79.8 & 80.6 & 0.609 & 0.670\\ \hline
			TBJE~\cite{ACL2020Challenge-JointTransformer} & 45.0 & 82.4 & - & - & -\\ \hline	
			Mu-Net~\cite{ACL2020-Multilogue-Net} & - & 82.1 & 80.0 & 0.590 & 0.50\\ \hline	
			MR~\cite{ACL2020-MultimodalRouting} & 51.6	& 81.7	& 81.8	& - & -\\ \hline					
			MulT~\cite{ACL2019-MultmodalTransformer} & 51.8 & 82.5 & 82.3 & 0.580 & 0.703\\ \hline
			\hline
			MCMulT (ours) & \textbf{52.4} & \textbf{83.1} & \textbf{82.8} & \textbf{0.582} & \textbf{0.706}\\ \hline
			\hline
			\multicolumn{6}{|c|}{CMU-MOSEI-Unaligned}\\ \hline
			CTC~\cite{ICML2016-CTC}+EF-LSTM & 46.3 & 76.1 & 75.9 & 0.680 & 0.585\\ \hline
			LF-LSTM & 48.8 & 77.5 & 78.2 & 0.624 & 0.656\\ \hline
			CTC+RAVEN~\cite{AAI2019-WordShift} & 45.5 & 75.4 & 75.7 & 0.664 & 0.599\\ \hline
			CTC+MCTN~\cite{AAAI2019-Translation-Modal} & 48.2 & 79.3 & 79.7 & 0.631 & 0.645\\ \hline
			MulT~\cite{ACL2019-MultmodalTransformer} & 50.7 & 81.6 & 81.6 & 0.591 & 0.694\\ \hline\hline
			MCMulT (ours) & \textbf{51.8} & \textbf{83.0} & \textbf{82.8} & \textbf{0.588} & \textbf{0.699}\\ \hline
		\end{tabular}
		}
	\caption{Results on (relatively large-scale) CMU-MOSEI with aligned and unaligned multimodal sequences.}
	\label{tab:result-CMU-MOSEI}

\end{table}

\subsection{Baselines}
Our MCMulT architecture is compared with early multimodal fusion LSTM (EF-LSTM), late multimodal fusion LSTM (LF-LSTM), Recurrent Attended Variation Embedding Network (RAVEN)~\cite{AAI2019-WordShift}, Multimodal Cyclic Translation Network (MCTN)~\cite{AAAI2019-Translation-Modal} and MulT~\cite{ACL2019-MultmodalTransformer}, where MulT achieved SOTA results on various multimodal unaligned sentiment recognition tasks. Following MulT, we apply connectionist temporal classification (CTC)~\cite{ICML2016-CTC} in methods (e.g. EF-LSTM, MCTN, RAVEN) which cannot be applied directly to the word-unaligned setting. In the experiment of the word-aligned setting, Recurrent Multistage Fusion Network (RMFN)~\cite{emnlp-2018-liang-multimodal}, Multimodal Factorization Model (MFM)~\cite{ICLR-2019-factorized-multimodal-representations}, Multilogue-Net (MU-Neet)~\cite{ACL2020-Multilogue-Net}, Multimodal Routing~\cite{ACL2020-MultimodalRouting}, Transformer-based joint-encoding (TBJE)~\cite{ACL2020Challenge-JointTransformer} and MulT methods are added to compare with MCMulT.

\begin{table*}[!h]
	\scriptsize
	\centering
	\vspace{5px}
	\setlength{\tabcolsep}{4mm}{
	\begin{tabular}	{|c|c|c|c|c|c|c|c|c|} 
		\hline
		Task & \multicolumn{2}{|c|}{Happy} & \multicolumn{2}{|c|}{Sad} & \multicolumn{2}{|c|}{Angry} & \multicolumn{2}{|c|}{Neutral}\\\hline
		Metric & $\text{Acc}^h$ & $\text{F1}^h$ & $\text{Acc}^h$ & $\text{F1}^h$ & $\text{Acc}^h$ & $\text{F1}^h$ & $\text{Acc}^h$ & $\text{F1}^h$\\ \hline
		\hline
		\multicolumn{9}{|c|}{IEMOCAP-Aligned}\\ \hline
		\hline
		EF-LSTM & 86.0 & 84.2 & 80.2 & 80.5 & 85.2 & 84.5 & 67.8 & 67.1\\ \hline
		LF-LSTM & 85.1 & 86.3 & 78.9 & 81.7 & 84.7 & 83.0 & 67.1 & 67.6\\ \hline
		RMFN~\cite{emnlp-2018-liang-multimodal} & 87.5 & 85.8 & 83.8 & 82.9 & 85.1 & 84.6 & 69.5 & 69.1\\ \hline
		MFM~\cite{ICLR-2019-factorized-multimodal-representations} & 90.2 & 85.8 & 88.4 & 86.1 & 87.5 & 86.7 & 72.1 & 68.1\\ \hline
		RAVEN~\cite{AAI2019-WordShift} & 87.3 & 85.8 & 83.4 & 83.1 & 87.3 & 86.7 & 69.7 & 69.3\\ \hline
		MCTN~\cite{AAAI2019-Translation-Modal} & 84.9 & 83.1 & 80.5 & 79.6 & 79.7 & 80.4 & 62.3 & 57.0\\ \hline
		MR~\cite{ACL2020-MultimodalRouting}	& 87.3 & 84.7 &	85.7 & 85.2	& 87.9	& 87.7	& 70.4	& 70.0\\ \hline
		MCMulT (ours) & \textbf{91.4} & \textbf{88.8} & \textbf{87.4} & \textbf{86.8} & \textbf{87.6} & \textbf{87.3} & \textbf{72.9} & \textbf{71.5}\\ \hline
		\hline
		\multicolumn{9}{|c|}{IEMOCAP-Unaligned}\\ \hline
		CTC~\cite{ICML2016-CTC}+EF-LSTM & 76.2 & 75.7 & 70.2 & 70.5 & 72.7 & 67.1 & 58.1 & 57.4\\ \hline
		LF-LSTM & 72.5 & 71.8 & 72.9 & 70.4 & 68.6 & 67.9 & 59.6 & 56.2\\ \hline
		CTC+RAVEN~\cite{AAI2019-WordShift} & 77.0 & 76.8 & 67.6 & 65.6 & 65.0 & 64.1 & 62.0 & 59.5\\ \hline
		CTC+MCTN~\cite{AAAI2019-Translation-Modal} & 80.5 & 77.5 & 72.0 & 71.7 & 64.9 & 65.6 & 49.4 & 49.3\\ \hline
		MulT~\cite{ACL2019-MultmodalTransformer} & 84.8 & 81.9 & 77.7 & 74.1 & 73.9 & 70.2 & 62.5 & 59.7\\ \hline\hline
		MCMulT (ours) & \textbf{86.0} & \textbf{83.2} & \textbf{78.8} & \textbf{75.3} & \textbf{74.6} & \textbf{71.1} & \textbf{62.7} & \textbf{60.37}\\ \hline
	\end{tabular}
	}
	\caption{Results on IEMOCAP with aligned and unaligned multimodal sequences.}
	\label{tab:result-IEMOCAP}
\end{table*}

\subsection{Quantitative Analysis}
\textbf{Word-Unaligned Experiments.} We compare our MCMulT model with prior approaches on three data datasets in the unaligned setting. The results are demonstrated in the bottom part of Table~\ref{tab:result-CMU-MOSI},\ref{tab:result-CMU-MOSEI}, \ref{tab:result-IEMOCAP}\footnote{The results of existing works in Table~\ref{tab:result-CMU-MOSI},~\ref{tab:result-CMU-MOSEI}, \ref{tab:result-IEMOCAP} are reported as Tsai et al.~\cite{ACL2019-MultmodalTransformer}.}. Our MCMulT achieves higher performance than the prior methods~\cite{AAI2019-WordShift,AAAI2019-Translation-Modal,ACL2019-MultmodalTransformer}, especially improves more than 1.0\% than the existing state-of-the-art MulT method on most $\text{Acc}$ and $\text{F1}$ attributes. Further analysis shows that MCMulT and MulT methods in word-aligned setting obtain slightly better performance than word-unaligned setting. However, for the word-aligned approaches (LF-LSTM, EF-LSTM, RAVEN and MCTN), the gaps between aligned and unaligned settings are more than 5\%-15\% on most attributes. This shows that MCMulT and MulT methods are more effective to tackle the asynchronous nature from multimodal sequences.

\textbf{Word-Aligned Experiments.} We evaluate our MCMulT model and the existing approaches~\cite{emnlp-2018-liang-multimodal,ICLR-2019-factorized-multimodal-representations,AAI2019-WordShift,AAAI2019-Translation-Modal,ACL2019-MultmodalTransformer,ACL-2018-bagher-zadeh-multimodal} in the case of word-aligned setting on the same datasets. The results are shown in the top part of Table~\ref{tab:result-CMU-MOSI},~\ref{tab:result-CMU-MOSEI},~\ref{tab:result-IEMOCAP}. MCMulT outperforms the other competitive approaches on different metrics on all attributes.

\textbf{Ablation Study.} We perform experiments with different variants of MCMulT: MulT with 7-layer, 10-layer and 12-layer structures, MCMulT-Dense, MCMulT-LocalDense and MCMulT-Global. As illustrated in Table 5, the architectures (MCMulT-LocalDense, MCMulT-Dense, MCMulT-Global) obtain worse performance than MCMulT. It shows that the appropriate interaction complexity of multi-scale mechanism obtains better performance. The architectures (MulT with 7-layer, 10-layer and 12-layer) obtain worse performance than 5-layer architecture, which shows increasing MulT complexity properly can not get better performance. Meanwhile, when MCMulT and MulT-12 have same network depth, MCMulT achieve better performance than MulT-12, which shows the effectiveness of multi-scale mechanism. Table~\ref{tab:result-variants},~\ref{tab:result-crossmodal},~\ref{tab:result-hyperparameters}. 

\begin{table}[!t]
	\scriptsize
	\centering
	\begin{tabular}	{|c|c|c|c|c|c|} 
		\hline
		Metric & $\text{Acc}^h_7$ & $\text{Acc}^h_2$ & $\text{F1}^h$ & $\text{MAE}^l$ & $\text{Corr}^h$\\ \hline
		MCMulT-Dense & 50.27 & 80.74 & 80.39 & 0.605 & 0.677\\ \hline
		MCMulT-LocalDense & 51.36 & 82.41 & 82.11 & 0.593 & 0.696\\ \hline
		MCMulT & 51.83 & 83 & 82.77 & 0.588 & 0.699\\ \hline
		MulT-5 & 50.70 & 81.60 & 81.60 & 0.591 & 0.694\\ \hline
		MulT-7 & 50.47 & 81.23 & 81.19 & 0.596 & 0.691\\ \hline
		MulT-10 & 48.64 & 78.89 & 78.76 & 0.619 & 0.672\\ \hline
		MulT-12 & 46.23	& 76.77	& 76.89	& 0.636	& 0.659\\ \hline
	\end{tabular}
	\caption{Results on  variants of MCMulT using CMU-MOSEI.}
	\label{tab:result-variants}
\end{table}
\begin{table}[t]
	\scriptsize
	\centering
	\begin{tabular}	{|c|c|c|c|c|c|} 
		\hline
		Metric & $\text{Acc}^h_7$ & $\text{Acc}^h_2$ & $\text{F1}^h$ & $\text{MAE}^l$ & $\text{Corr}^h$\\ \hline
		text only & 46.80 & 77.70 & 78.60 & 0.654 & 0.633\\ \hline
		vision only & 43.70 & 66.90 & 70.20 & 0.754 & 0.347\\
		\hline
		audio only & 42.10 & 65.20 & 68.90 & 0.761 & 0.313\\
		\hline\hline
		MulT (V,A-\textgreater T) & 50.5 & 80.1 & 80.4 & 0.605 & 0.670\\ \hline
		MulT (T,A-\textgreater V) & 48.2 & 79.7 & 80.2 & 0.611 & 0.648\\ \hline
		MulT (T,V-\textgreater A) & 47.5 & 79.2 & 79.7 & 0.620 & 0.653\\ \hline\hline
		MCMulT (V,A-\textgreater T) & 50.60 & 80.80 & 80.90 & 0.601 & 0.677\\ \hline
		MCMulT (T,A-\textgreater V) & 49.10 & 80.10 & 80.50 & 0.607 & 0.657\\ \hline
		MCMulT (T,V-\textgreater A) & 47.80 & 79.80 & 80.50 & 0.613 & 0.653\\ \hline
		MCMulT & 51.8 & 83.00 & 82.8 & 0.588 & 0.699\\ \hline
	\end{tabular}
	\caption{Results on the benefit of MCMulT’s crossmodal transformers using CMU-MOSEI.}
	\label{tab:result-crossmodal}
		\vspace{-10px}
\end{table}

\begin{table}[t]
	\scriptsize
	\centering
	\begin{tabular}	{|c|c|c|c|c|c|} 
		\hline
		Metric & $\text{Acc}^h_7$ & $\text{Acc}^h_2$ & $\text{F1}^h$ & $\text{MAE}^l$ & $\text{Corr}^h$\\ \hline
		B=1 & 47.60 & 78.20 & 78.10 & 0.655 & 0.633\\ \hline
		B=2 & 50.30 & 80.70 & 80.60 & 0.603 & 0.675\\ \hline
		B=3 & 51.61 & 82.31 & 82.17 & 0.593 & 0.693\\ \hline
		B=4 & 51.83 & 83.0 & 82.77 & 0.588 & 0.699\\ \hline
		B=5 & 41.26 & 82.13 & 82.81 & 0.591 & 0.695\\
		\hline\hline
		L=1 & 50.60 & 80.90 & 80.40 & 0.603 & 0.673\\ \hline
		L=2 & 51 & 81.73 & 80.90 & 0.596 & 0.688\\ \hline
		L=3 & 51.83 & 83 & 82.77 & 0.588 & 0.699\\ \hline
		L=4 & 51.37 & 82.22 & 82.04 & 0.592 & 0.694\\ \hline
	\end{tabular}
	\caption{Results on different value of hyperparameters of MCMulT. B and L indicate the block number and layer number of each block, respectively.}
	\label{tab:result-hyperparameters}

\end{table}


1) Single Modality and Single Target Modality. We have conducted comparative experiments of single modality (only considering textual, visual or acoustic modality) and single target modality ($\tiny[A,V\rightarrow L], [L,V\rightarrow A], [L,A\rightarrow V]$). The results indicate that the MCMulT architecture is efficient and its performance is better than the MulT architecture in the same setting. The experimental results are shown in Table~\ref{tab:result-crossmodal}. In experiments of single modality, the performance of text modality is much better than that of visual and audio modalities, which is consistent with the conclusion of previous work. In the single-target modality experiments, the performances ($\text{Acc}2$ and F1) of visual and audio modalities based on MCMulT are improved by 10\%-15\% (same as the phenomenon of MulT) compared with the corresponding single modality, and  the performance of the text modality is also increased by about 3\%. Furthermore, the performance of single-target modality obtained by MCMulT is better than MulT on all attributes. This shows that our MCMulT models crossmodality interactions more effectively than MulT.

2) MCMulT Hyperparameters.
We explore the influence of the number of MCTB and CT layers in the MCMulT architecture. The experimental results are shown in Table~\ref{tab:result-hyperparameters}. The performance is improved as the number of MCTB increases from 1 to 4, and 4 is the best parameter. CT follows the similar process as MCTB, when the number of CT is 3, the best performance is achieved. We hypothesize such phenomenon occurs because of the learning ability of MCMulT and the dataset scale.

\section{Conclusion}
In this paper, we proposes a MCMulT architecture, which utilizes multi-scale and cooperative crossmodal interactions from unaligned multimodal sequences to solve sentiment analysis task. Specifically, to improve the robustness of a transformer, during the crossmodal interaction, we use multi-scale representations that contain different levels of semantic information. Additionally, we have also designed an  efficient crossmodal interaction method in terms of computational cost.
The variants of MCMulT are further discussed. The experimental results clearly demonstrate that our approach outperforms the state-of-the-art works on not only on unaligned datasets but also on aligned multimodal datasets. 
In the future work, we will study finer architecture of MCMulT, e.g, layer number tuning individually for each modality to achieve better performance and reduce computation cost. Besides, we plan to extend MCMulT to more multimodal applications with large-scale datasets, e.g., VQA, image-text matching and crossmodal pretraining tasks.

\bibliography{custom}

\end{document}


\maketitle

\begin{figure*}[!h]
	\begin{center}
		\includegraphics[width=1.0\linewidth]{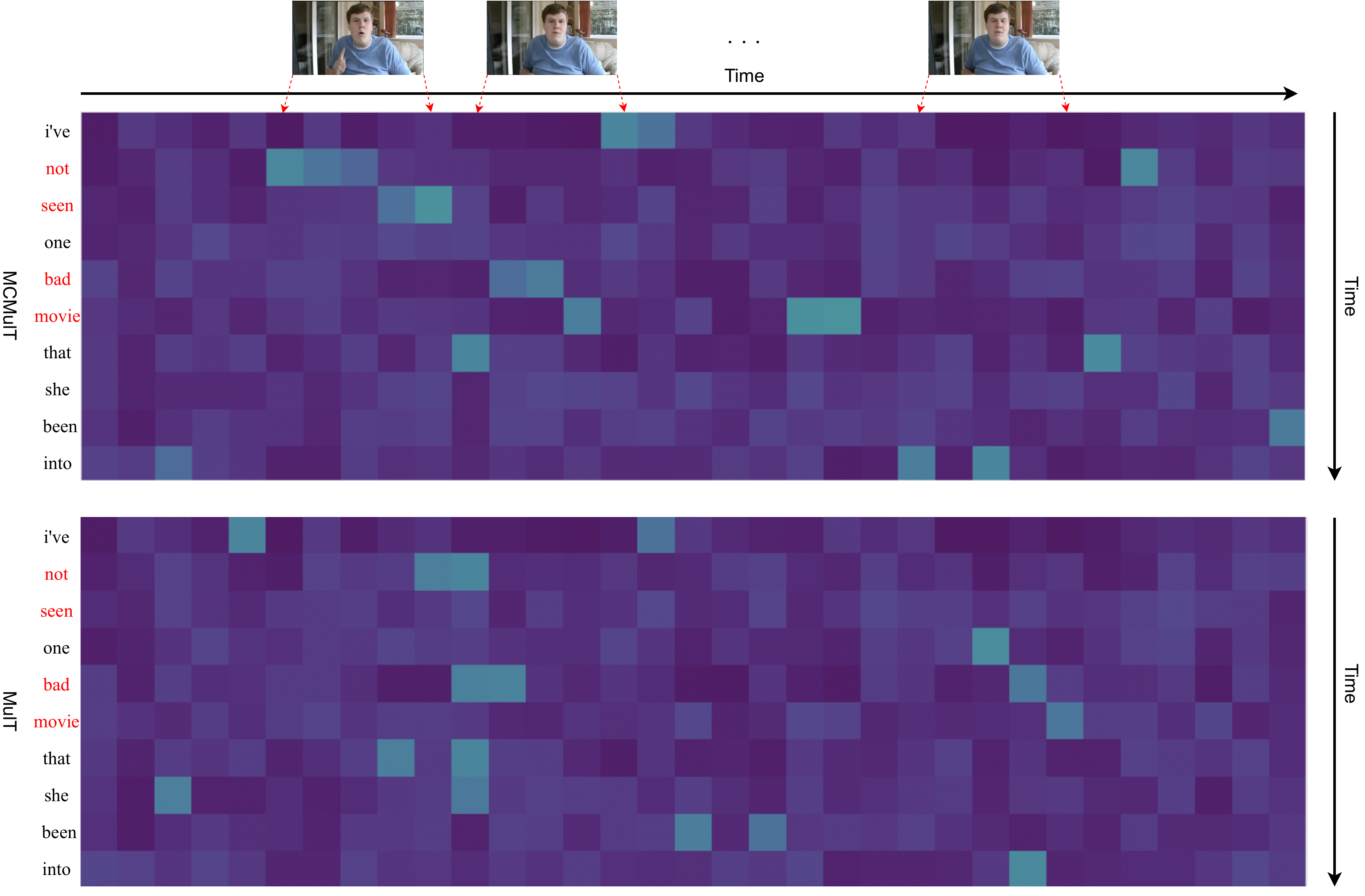}
		\caption{Illustration of attention map from MCMulT and MulT.}
		\label{fig:Visualization}
	\end{center}
\end{figure*}

\section{Appendix A}
\subsection{Datasets}
The datasets in our experiments are used as did in paper ~\cite{ACL2019-MultmodalTransformer}.
The division setting is provided by the respective benchmarks and also utilized by the state-of-the-art methods~\cite{ACL2019-MultmodalTransformer}. We strictly follow their settings.
Tsai~\cite{ACL2019-MultmodalTransformer} provides the link to the downloadable datasets as below. 
Data files (containing processed MOSI, MOSEI and IEMOCAP datasets) can be downloaded from ~\cite{Dataset-Url}. To get the meta information and the raw data, please refer to the SDK for these datasets from ~\cite{Dataset-sdk}.

\subsection{Experimental Details}
The number of scale layers is a hyperparameter for each modality. If the hierarchy of a certain modality is relatively simple, it is reasonable to have fewer blocks for that modality. Fine-tuning hypermeters may also likely improve performance because of avoiding overfitting. We set the same number of scale layers for all modalities for simplicity in the experiments. This setting makes the paper easy to understand and follow.

Cross-Entropy loss is used to train MCMulT. The epoch number and batch size are chosen as 100 and 64 respectively. A 0.2 dropout rate is picked for all attention layers and a 0.1 dropout rate is picked for all full-connection layers. Furthermore, a gradient clipping value of 0.8 is applied, and our architecture's trainable parameters are optimized using Adam with 1e-3 learning rate. All models are implemented using PyTorch framework.

\section{Appendix B}
\subsection{Example Visualization}

Figure~\ref{fig:Visualization} illustrates one example on crossmodal attention maps between vision and text modality that are from high-level layer of our MCMulT and MulT. Some important words are colorized as red, and green color in attention map indicates higher attention score than blue color in Figure~\ref{fig:Visualization}. It shows that the word ``not'' and ``seen'' are coordinately aligned with neighboring frames of facial expression in MCMulT, the phrase ``bad movie'' has the similar result. By contrast, the alignment for ``not'' and ``seen'' are irrelevant in MulT.

\bibliography{custom}